\documentclass[runningheads]{llncs}
\usepackage{graphicx}
\usepackage{float}
\usepackage{subfig}
\usepackage{amsmath}
\usepackage{amssymb}
\usepackage{xcolor}
\usepackage{acronym}
\acrodef{DL}[DL]{deep learning}
\acrodef{CNN}[CNN]{convolutional neural network}
\acrodef{FPGA}[FPGA]{field-programmable gate array}
\acrodef{BNN}[BNN]{binarized neural network}
\acrodef{FPS}[FPS]{frames per second}
\acrodef{IoU}[IoU]{intersection over union}
\acrodef{UAV}[UAV]{unmanned aerial vehicle}
\acrodef{ASIC}[ASIC]{application-specific integrated circuit}
\acrodef{AI}[AI]{artificial intelligence}
\acrodef{NIR}[NIR]{near-infrared}
\acrodef{CAD}[CAD]{computer-aided design}
\acrodef{IR}[IR]{infrared}
\acrodef{PL}[PL]{programmable logic}
\acrodef{PS}[PS]{processing system}
\acrodef{SNN}[SNN]{spiking neural networks}
\acrodef{MCC}[MCC]{Matthews correlation coefficient}
\acrodef{HAF}[HAF]{Hafiane quality index}

\begin{document}
\title{An FPGA smart camera implementation of segmentation models for drone wildfire imagery}
\titlerunning{FPGA and smart camera implementation for wildfire segmentation}
%

\author{
Eduardo Guarduño-Martinez\inst{1} \and
Jorge Ciprian Sanchez\inst{2} \and
Gerardo Valente Vazquez-Garcia  \inst{3} \and 
Gerardo Rodriguez-Hernandez \inst{1} \and Adriana Palacios-Rosas \inst{4} \and Lucile Rossi-Tisson \inst{5} \and
Gilberto Ochoa-Ruiz\inst{1}}

\authorrunning{E. Guarduño-Martinez et al.}
%
\institute{Tecnologico de Monterrey, School of Engineering and Sciences, Mexico 
\and
Hasso-Plattner Institute, University of Postdam, Germany
\and
Maestria en Cs. Computacionales, Universidad Autonoma de Guadalajara, Mexico
\and Universidad de las Americas Puebla, Department of Chemical, Food and Environmental Engineering, Mexico \and Universit\`a di Corsica, Laboratoire Sciences Pour l’Environnement,\\ Campus Grimaldi – BP, Corti, France}

\maketitle              

\vspace{-4mm}
\begin{abstract}
Wildfires represent one of the most relevant natural disasters worldwide, due to their impact on various societal and environmental levels. Thus, a significant amount of research has been carried out to investigate and apply computer vision techniques to address this problem. One of the most promising approaches for wildfire fighting is the use of drones equipped with visible and infrared cameras for the detection, monitoring, and fire spread assessment in a remote manner but in close proximity to the affected areas. However, implementing effective computer vision algorithms on board is often prohibitive since deploying full-precision deep learning models running on GPU is not a viable option, due to their high power consumption and the limited payload a drone can handle. Thus, in this work, we posit that smart cameras, based on low-power consumption \acp{FPGA}, in tandem with \acp{BNN}, represent a cost-effective alternative for implementing onboard computing on the edge. Herein we present the implementation of a segmentation model applied to the Corsican Fire Database. We optimized an existing U-Net model for such a task and ported the model to an edge device (a Xilinx Ultra96-v2 \ac{FPGA}). By pruning and quantizing the original model, we reduce the number of parameters by 90\%. Furthermore, additional optimizations enabled us to increase the throughput of the original model from 8 \ac{FPS}  to 33.63 \ac{FPS} without loss in the segmentation performance: our model obtained 0.912 in \ac{MCC}, 0.915 in F1 score and 0.870 in \ac{HAF}, and comparable qualitative segmentation results when contrasted to the original full-precision model.
The final model was integrated into a low-cost \ac{FPGA}, which was used to implement a neural network accelerator.

\end{abstract}
\section{Introduction}
\label{sec:introduction}

A wildfire is an exceptional or extraordinary free-burning vegetation fire that may have been started maliciously, accidentally, or through natural means that could significantly affect the global carbon cycle by releasing large amounts of C02 into the atmosphere. It has profound economic effects on people, communities, and countries, produces smoke that is harmful to health, devastates wildlife, and negatively impacts bodies of water \cite{UnitedNations}. The three main categories of remote sensing for wildfire monitoring and detection systems are ground-based systems, manned aerial vehicle-based systems, and satellite-based systems. However, they present the following technological and practical problems: ground-based have limited surveillance ranges. Satellite-based have problems when planning routes, their spatial resolution may be low, and the information transmission may be delayed. Manned aerial vehicle-based systems are expensive and potentially dangerous due to hazardous environments and human error. \Acp{UAV} provide a mobile and low-cost solution using computer vision-based remote sensing systems that can perform long-time, monotonous, and repetitive tasks \cite{YuanSurvey}. Drones, in particular, represent an excellent opportunity due to their easy deployment. However, the ability to implement these fire detection systems, based on \ac{DL}, is limited by the maximum payload of the drone and the high power consumption. 

In this paper, we posit that a \ac{CNN} can be implemented on a hardware accelerator that can be embedded as part of a smart camera and installed on a drone for the detection of wildfires. A review of the literature on hardware implementation for various \ac{AI} algorithms was published by Talib et al. \cite{Talib2021} reviewing 169 different research reports published between 2009 and 2019, which focus on the implementation of hardware accelerators by using \acp{ASIC}, \acp{FPGA}, or GPUs. They found that most implementations were based on \acp{FPGA}, focusing mainly on the acceleration of \acp{CNN} for object detection, letting the GPU-based implementations in second place.

Due to the diversity of applications, \ac{AI} models such as \acp{CNN} need to meet various performance requirements for drones and autonomous vehicles, with the essential demands of low latency, low weight overhead, long-term battery autonomy, and low power consumption being the most pressing requirements. The complexity of the tasks that \acp{CNN} must perform continues to increase as models evolve. As a result, deeper networks are designed in exchange for higher computational and memory demands. In this context, the reconfiguration capabilities of \acp{FPGA} enable the creation of \ac{CNN} hardware implementations that are high-performance, low-power, and configurable to fit system demands \cite{Venieris}. A smart camera is an embedded system for computer vision applications that has attracted great interest in various application domains, as it offers image capture and image processing capabilities in a compact system \cite{sota_shi_2009}. This paper describes the methodology, implementation, design cycle, and experimental protocol of porting a modified U-Net model into a Xilinx Ultra96-V2 \ac{FPGA} for the wildfire semantic segmentation task for the smart camera system.

The rest of the paper is organized as follows: Section \ref{sec:state_of_the_art} discusses recent works applying computer vision models for wildfire segmentation, highlighting their strengths and limitations; the second part of the section discusses related works regarding smart camera implementations in order to better contextualize our work. Section \ref{sec:proposed_method} details our contribution, discussing in detail the proposed model, the dataset used for evaluating our models, and the design flow followed for optimizing the model and testing it in the target embedded \ac{FPGA} board. Section \ref{sec:results_discussion} discusses the results of our optimization process and provides a quantitative and qualitative comparison between full precision and the \ac{BNN} model. Finally, Section \ref{sec:conclusions} concludes the paper and discusses future areas of research.

\section{State-of-the-art}
\label{sec:state_of_the_art}

\subsection{Segmentation Models for Wildfire Detection and Characterization}
\label{subsec:seg_models_wildfire_detection}

Detecting a wildfire by categorizing each pixel in an infrared image is a semantic segmentation problem; therefore, for this task, \ac{AI} models have been used, such as fully convolutional networks as well as the U-Net model proposed by Ronnenberger et al. in 2015 \cite{unet_2015}, which allow precise segmentation with few training images. For the specific task of fire segmentation, artificial intelligence models have already been implemented to solve this problem with visible images of fire \cite{Akhloufi}, the fusion of visible and infrared images of fire \cite{Jorge}, and visible images of fire and smoke \cite{Perrolas}. For instance, Akhloufi et al. \cite{Akhloufi} proposed Deep-Fire, a semantic segmentation model based on the U-Net architecture. The authors trained and evaluated their model using the Corsican Fire Database \cite{Corsican}. With an F1 score ranging from 64.2\% to 99\% on the test set, Akhloufi et al. claimed successful results using the Dice similarity coefficient as the loss function for the model.

Ciprián-Sánchez et al. \cite{Jorge} evaluated thirty-six different \ac{DL} combinations of the U-Net-based Akhloufi architecture \cite{Akhloufi}, the FusionNet-based Choi architecture \cite{Choi2021}, and the VGG16-based Frizzi architecture \cite{Frizzi2021}, the Dice \cite{Ma2020}, Focal Tversky \cite{Tversky}
, Unified Focal \cite{Yeung2021} losses, and the visible and \ac{NIR} images of the Corsican Fire Database \cite{Corsican} and fused visible-\ac{NIR} images produced by the methods by Li et al. \cite{HuiLi} and Ciprián-Sánchez et al. \cite{Jorge_2}. After evaluating these models, the combination with the best results was Akhloufi + Dice + visible with a 0.9323 F1 score, also known as the Dice coefficient.

Although these works have highlighted the potential of using \ac{AI} in this domain, many of these algorithms are incapable of operating in real-time,  as they inherently suffer from very high inference times and are prohibitive as they require many computing resources, which impedes their usability on drone missions and thus we posit that new paradigms are needed for their successful deployment, particularly in terms of inference time (\ac{FPS}) and power consumption.

\subsection{Smart Camera Implementations for Computer Vision}
\label{subsec:smart_camera_implementations}

Smart cameras are devices that process, analyze, and extract data from the images they capture. Different video processing algorithms are used for the extraction. 

Smart cameras have been employed in a variety of applications, including human gesture recognition \cite{sota_wolf_2002}, surveillance \cite{sota_bramberger_2006}, smart traffic signal optimization systems \cite{sota_tchuitcheu_2020}, and a fire detection system based on conventional image processing methods \cite{sota_gomes_2014}. We propose a \ac{DL} implementation capable of performing a precise segmentation that can be used as a first step in wildfire characterization and risk assessment systems.


\Acp{FPGA} are excellent choices for creating smart cameras because they offer significant processing capabilities while maintaining a low power consumption, which makes them good candidates for particular edge tasks creating efficient hardware accelerators capable of high throughput \cite{Venieris}, and maintaining a high degree of flexibility and reconfigurability. 

The disadvantage of \acp{FPGA} is that developers need to be skilled in hardware design to accomplish these goals. The design process frequently takes longer with \acp{FPGA} than with CPU and GPU systems. To address such issues, \ac{FPGA} vendors and other academic and industrial tool developers have introduced several \ac{CAD} tools for training and optimizing \ac{DL} models and mapping such models into the reconfigurable fabric.

Convolutional neural networks provide high-accuracy results for computer vision tasks, and their applications could be benefited from being implemented in edge devices such as \acp{FPGA}. Still, for applications such as smart cameras, limited use of hardware resources and power requirements are of the utmost importance. Therefore, to implement models that generally require a large number of computational resources, large storage capabilities for the model parameters, and the use of high-energy-consuming hardware \cite{sota_vestias_2019,Venieris}, such as GPUs, it is necessary to use model optimization techniques such as pruning and quantization \cite{sota_berthelier_2021} for the compression of the model, to implement it in devices such as an \ac{FPGA} while achieving high inference speed.


\section{Proposed Method}
\label{sec:proposed_method}

The implementation of a \ac{BNN} for the segmentation of wildfire images was done using the Xilinx tool Vitis AI because each operation of the model is mapped into a hardware-accelerated microinstruction, in which a series of sequential micro-instructions can represent the whole DL model, while a scheduler is in charge of managing the hardware calls and data flow. This enables the customization of the HW accelerator while considering the resources of the \ac{FPGA}. In the particular context of our application, Vitis AI is indeed the best choice as we target a small \ac{FPGA} device (Xilinx Ultra96-V2) for deep embedded image processing. 

\subsection{General Overview of the Optimization Approach}
\label{subsec:general_overview}

\begin{figure}[t]
  \centering
  \includegraphics[scale=0.22]{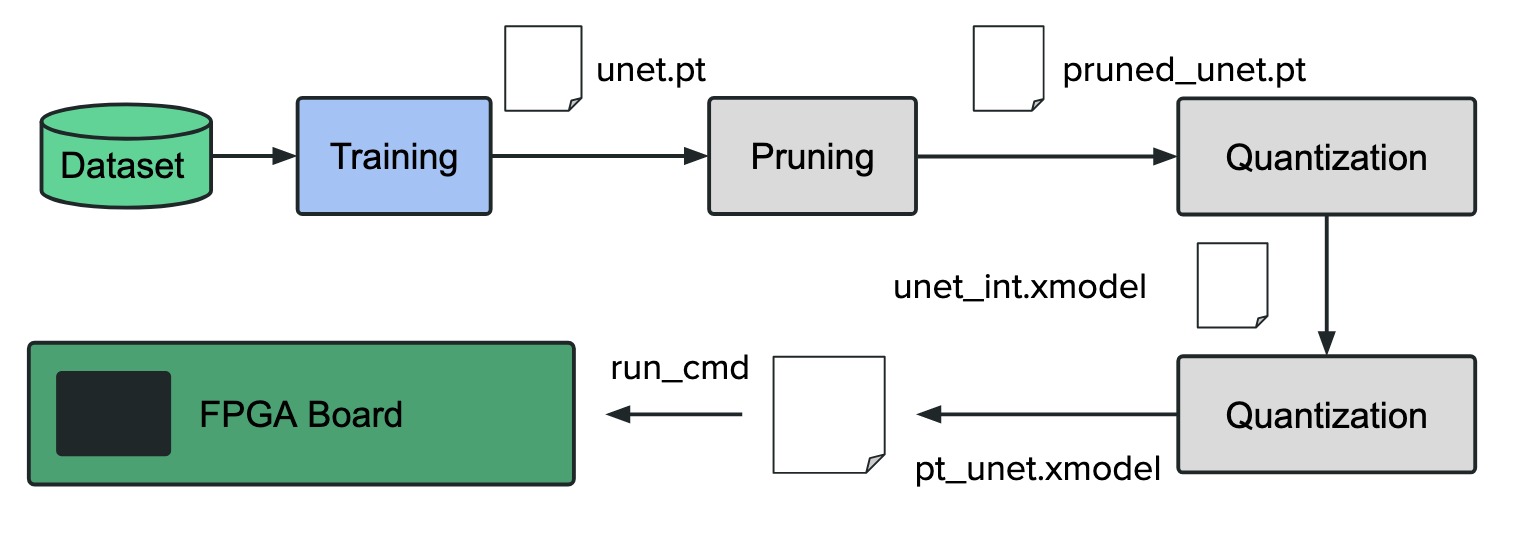}
  \caption{General overview of the Pytorch flow for Vitis AI. This flow allows us to optimize a given full precision model and target an embedded device such as an FPGA, consuming less power while attaining a higher throughput in terms of processed \ac{FPS}.}
  \label{fig:Pytorch flow}
\end{figure}

Fig. \ref{fig:Pytorch flow} depicts the Vitis AI Pytorch flow followed in this paper. The design process begins by training a segmentation with \ac{NIR} images from the Corsican Fire Database \cite{Corsican} and their corresponding ground truths for fire region segmentation. Subsequently, a  pruning process to reduce the number of filters in the convolution layers using the Pytorch framework is performed. Then, both the original and the pruned models are saved in pt files. The next module is in charge of changing the numerical representation of the \ac{DL} model by performing an 8-bit quantization using the Vitis AI quantizer module, producing an xmodel file. Finally, the quantized model is compiled, producing an xmodel file containing all the instructions needed by the DPU to execute the model.

After the model has been compiled, it can be loaded on the target \ac{FPGA} board and tested. In our work, this model is a U-Net model modified to accommodate the needs of our application. The rest of this section will detail the implementation of such an optimized segmentation model.

\subsection{Dataset: Corsican Fire Database}
\label{subsec:dataset}

\begin{figure}[H]
  \centering
  \includegraphics[scale=0.27]{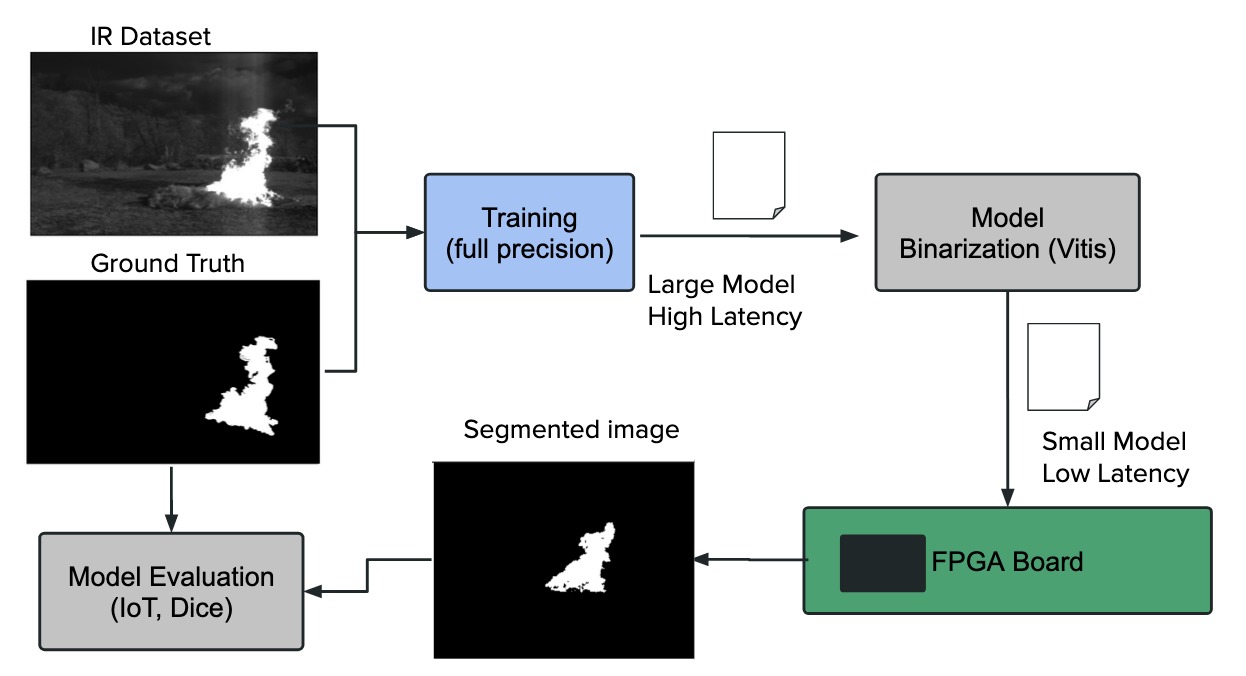}
  \caption{Overall implementation flow for \ac{FPGA}-based systems based on Vitis AI}
  \label{fig:Dataset}
\end{figure}

In this paper, we employ the \ac{NIR} images from the Corsican Fire Database, first introduced by Toulouse et al. \cite{Corsican}. For fire
region segmentation tasks, this dataset includes 640 pairs of visible and \ac{NIR} fire images along with the matching ground truths created manually by experts in the field.
A representative \ac{NIR} image from the Corsican Fire Database is shown in the top left corner of Fig. \ref{fig:Dataset}, along with its corresponding ground truth.

\subsection{Segmentation Model Training}
\label{subsec:seg_model_training}

\begin{figure}[t]
  \centering
  \includegraphics[scale=0.68]{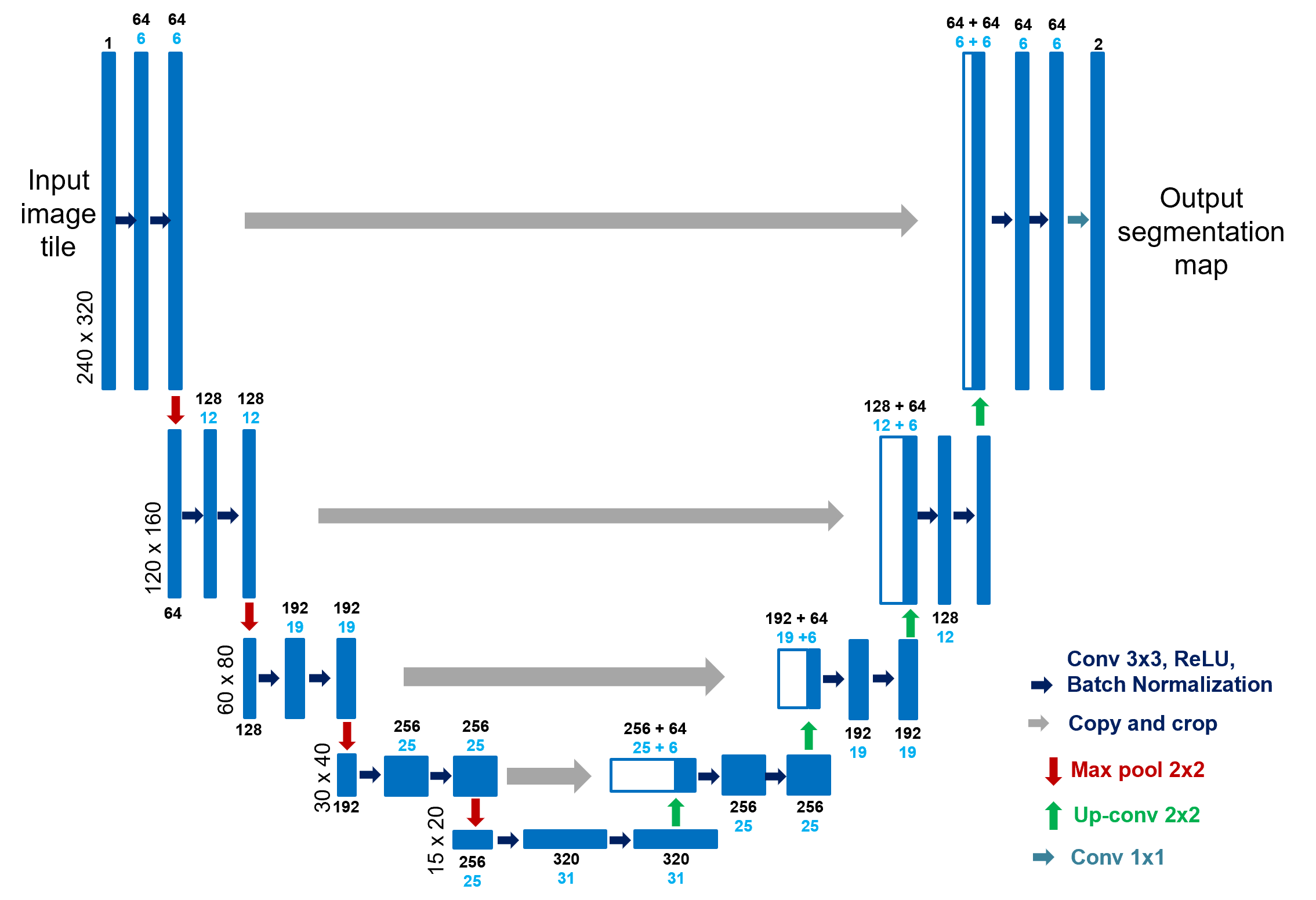}
  \caption{Proposed architecture. The original U-Net architecture has been extended by introducing batch normalization layers and fewer filters in the deepest layers to reduce training
and inference times. The numbers on black on the top of the blocks represent the original filter sizes, whereas the blue one (below) represents the filter size after the optimization process.}
  \label{fig:Proposed_architecture}
\end{figure}

The proposed architecture for this paper is a modified version of a U-Net model \cite{unet_2015} with the number of filters from the deepest layers reduced to reduce training and inference times. Furthermore, we add batch normalization layers \cite{Batch_Norm} after every convolutional layer. The final architecture is shown in Fig. \ref{fig:Proposed_architecture}; the numbers in black are the number of filters before pruning, and the numbers in blue are the number of filters after pruning. 

As depicted in Fig. \ref{fig:Dataset}, every image in the training set was resized to a width of 320 and a height of 240 pixels for training. For the training of the proposed model, the dataset was divided into 80\% for training and 20\% for testing. The model was trained with a learning rate of 0.0001 for 350 epochs with a batch size of 5 using cross-entropy loss and Adam optimizer.


\subsection{Optimization}
\label{subsec:optimization}

\subsubsection{Pruning\\}
\label{subsubsec:prunning}

\begin{figure}[H]
  \centering
  \includegraphics[scale=0.33]{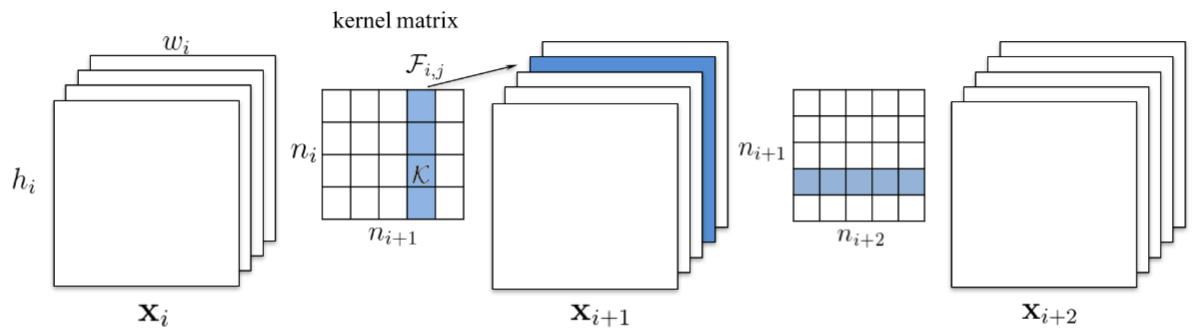}
  \caption{When a filter is pruned, the matching feature map and associated kernels in the following layer are removed. Retrieved from: Hao Li et al. \cite{HaoLi2016}.}
  \label{fig:Pruning}
\end{figure}

\begin{figure}[H]
  \centering
  \includegraphics[scale=0.14]{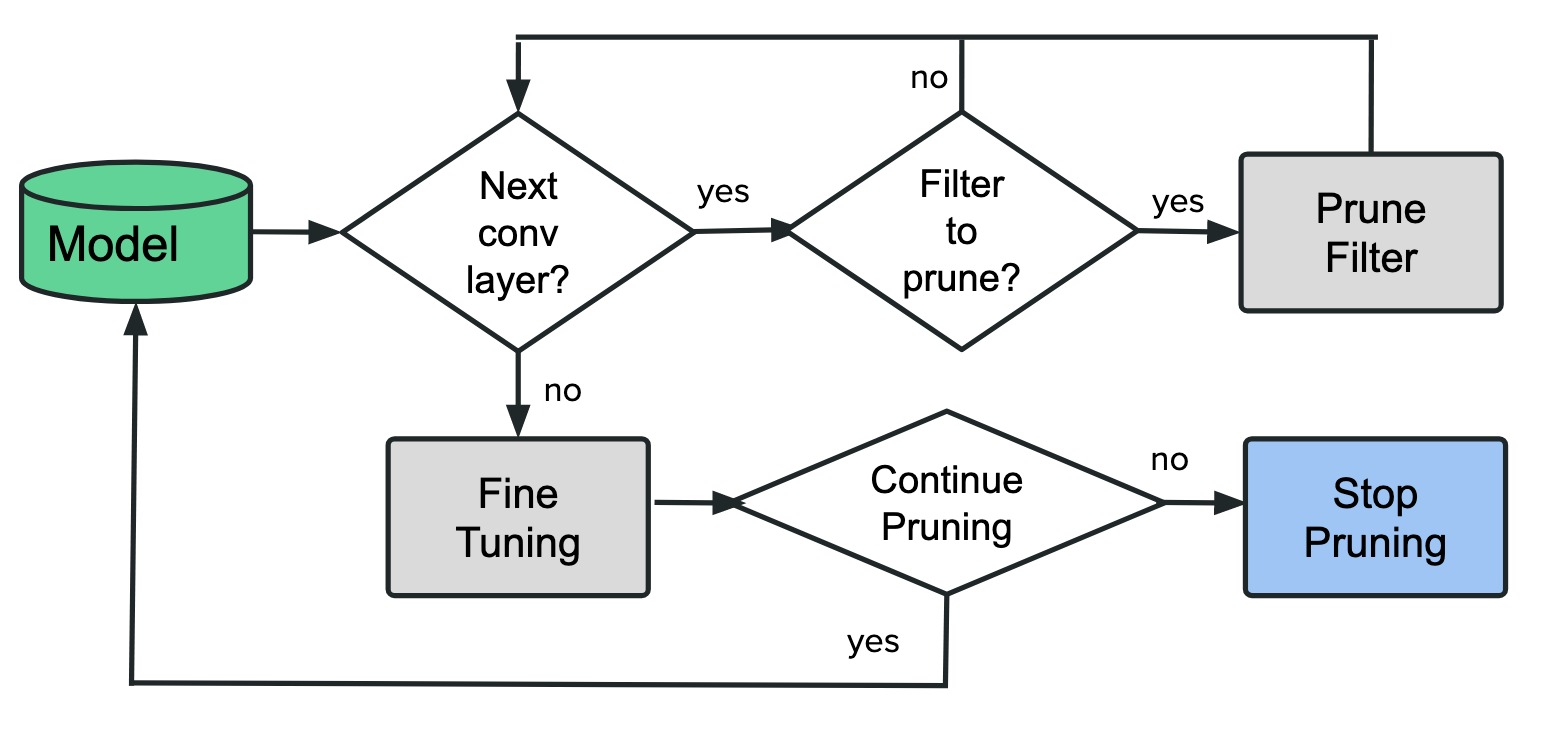}
  \caption{Schematic flow for optimizing a model in Vitis AI.}
  \label{fig:Pruning_process}
\end{figure}

The pruning method (contained in the binarization block of Fig. \ref{fig:Dataset}) employed in the present paper is based on the work of \cite{HaoLi2016} in which, as shown in Fig. \ref{fig:Pruning}, when a filter is pruned, the corresponding feature map is removed and the kernels of the input feature maps for the next convolution that correspond to the output feature maps of the pruned filters are also removed. Fig. \ref{fig:Pruning_process} briefly explains the pruning process used for this paper, with which it was possible to reduce the number of filters in each convolutional layer by approximately 90\%. In Fig. \ref{fig:Proposed_architecture}, we can see the final architecture of the model, the numbers in blue being the number of filters after the pruning process.

\subsubsection{Quantization\\}
\label{subsubsec:quantization}

The model was quantized using the Vitis AI quantizer module; resulting in a \ac{CNN} model with all its values represented with only 8 bits. That is, the floating-point checkpoint model in coverted into a fixed-point integer checkpoint. After confirming there was no significant degradation in the model's performance, the quantized model was compiled with the Vitis compiler, which creates a \textit{xmodel file} with all the instructions required by the DPU to execute the model.

\subsection{Proposed \ac{FPGA}-based Smart Camera System}
\label{subsec:fpga_smart_camera_system}

\begin{figure}[t]
  \centering
  \includegraphics[scale=0.26]{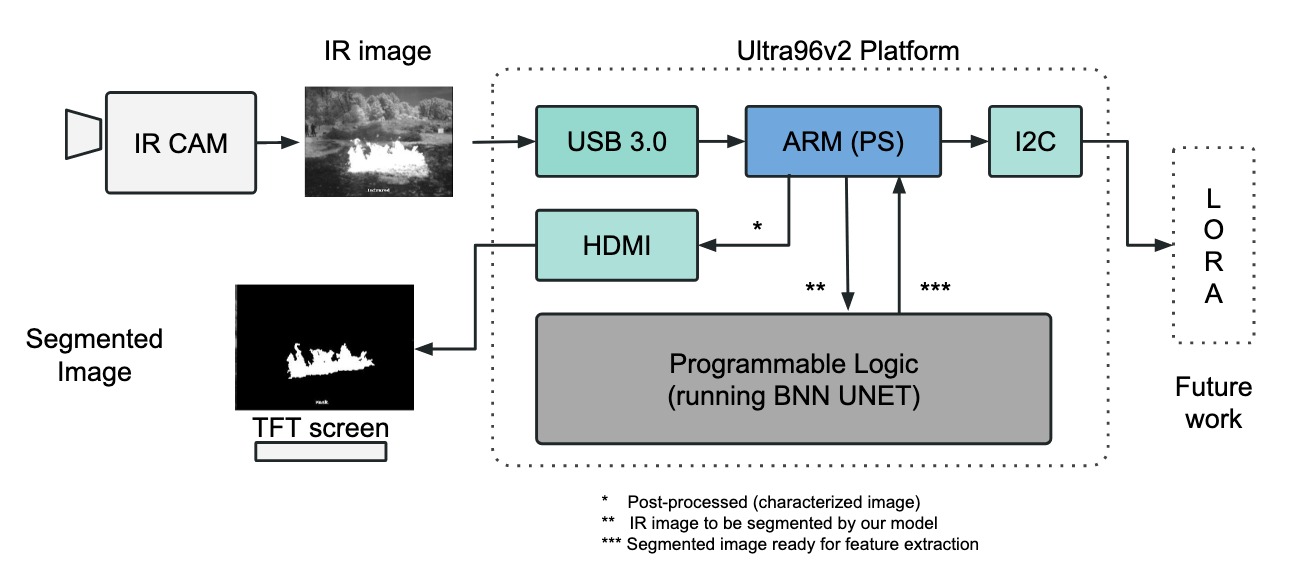}
  \caption{Proposed solution model for implementing a smart camera for wildfire detection. Our current implementation processes images from external memory or an \ac{IR} camera; communication capabilities have not yet been implemented.}
  \label{fig:Arch_general}
\end{figure}

Fig. \ref{fig:Arch_general} shows the system implementation for the smart camera solution of wildfire detection. The \ac{PS} controls every step of the application's life cycle, including retrieving images from the camera, feeding them to the \ac{PL} section of the SoC (hardware accelerator implementing the proposed model), and processing the segmented image.

An \ac{IR} camera is attached to the Ultra96 board using a USB port in the SoC. The \ac{PS} block (an ARM processor) processes the input picture before feeding it to the \ac{PL} section, which runs the binarized U-Net model mapped into the reconfigurable fabric. The image is processed and then passed back into the \ac{PS} block for feature extraction. If a complete IIoT solution is implemented, these features may be used for viewing on a TFT screen or communicated via a communication protocol (i.e., LORA) to a cloud. These capabilities are not yet implemented here and are left for future work. In order to make the picture more straightforward, the AXI connection, which is not illustrated here, is used for all communication between the \ac{PS}, \ac{PL}, and peripherals.

\begin{figure}[H]
  \centering
  \includegraphics[scale=0.28]{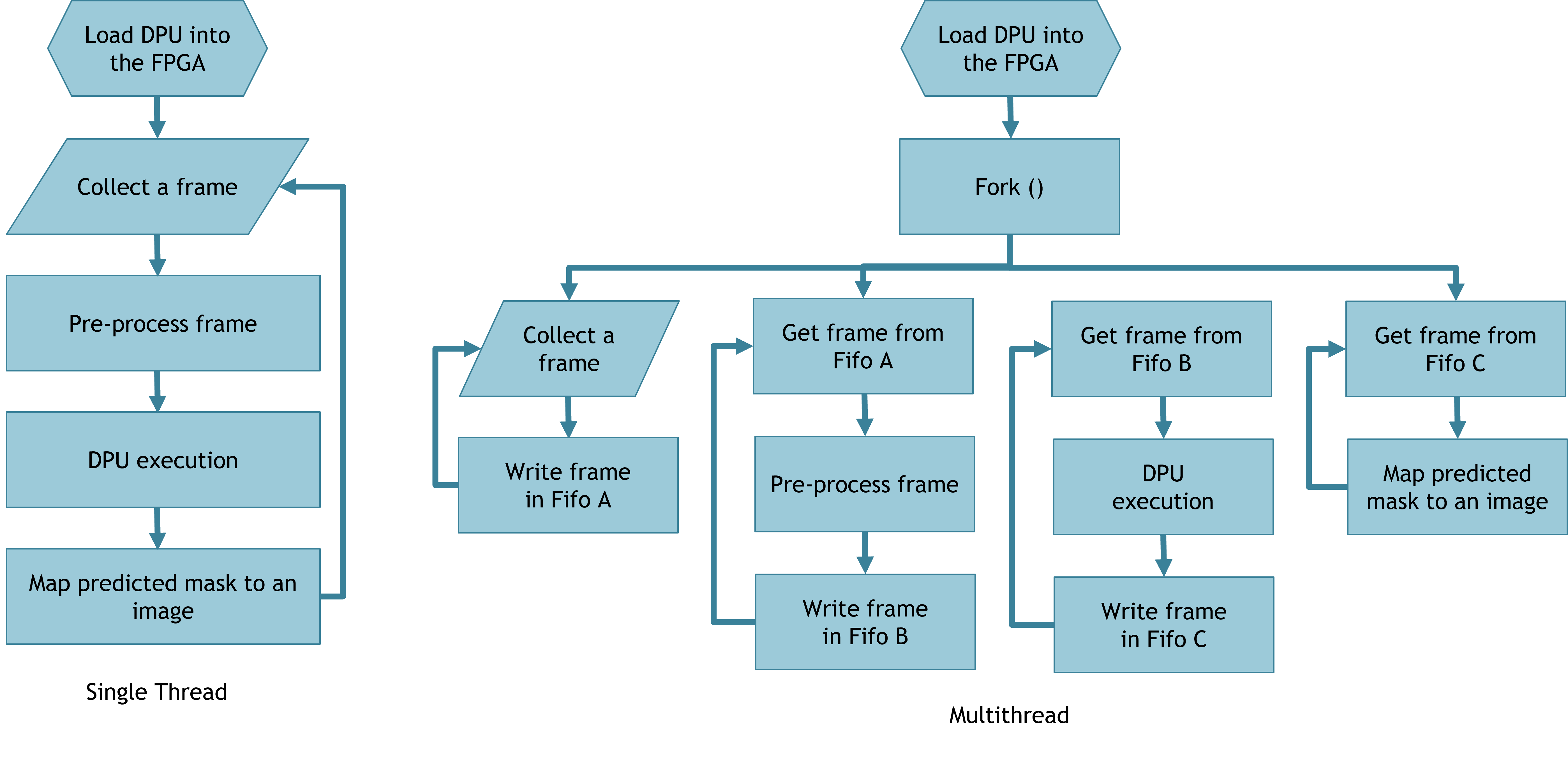}
  \caption{Flow chart for single and multi-threading inference approaches.}
  \label{fig:Flow_chart}
\end{figure}

In our experiments, the overall performance of the model implemented using single-thread execution was not satisfactory, as we obtained only a throughput of 15.77 \ac{FPS}, even after the pruning and quantization of the model. Therefore, we explored the use of a multi-thread approach supported by the Ultra89-v2 board. The use of this functionality enabled us to attain a higher performance. The main limitation of the single-threaded approach is the bottleneck introduced by the DPU when performing inference in the FPFA, as it introduces a significant latency. This problem arises from the use of queues for exchanging information among the different threads.  In Fig. \ref{fig:Flow_chart}, we provide a flow chart comparing both software implementations.

\section{Results and Discussion}
\label{sec:results_discussion}

In the subsequent section, we will discuss the results obtained from implementing the U-Net model for segmenting images of the Corsican Fire Database, comparing both the original full-precision model and the optimized model running on the \ac{FPGA} platform. We will also compare our results with previous works in the state-of-the-art based on a number of metrics used in the literature, which will be described in the next subsection. Then, quantitative and qualitative results will be provided, based on these metrics, followed by a discussion of the obtained results.

\subsection{Comparison Metrics}
\label{subsec:results_metrics}

\subsubsection{Matthews Correlation Coefficient}
\label{ch_03:mcc}

First proposed by Matthews \cite{Matthews75}, it measures the correlation of the true classes with their predicted labels \cite{Chicco21}. The \ac{MCC} represents the geometric mean of the regression coefficient and its dual, and is defined as follows \cite{Toulouse15}: 

\begin{equation}
\label{eq:mcc}
    MCC = \frac{(TP*TN)-(FP*FN)}{\sqrt{(TN+FN)(TN+FP)(TP+FN)(TP+FP)}},
\end{equation}

where $TP$ is the number of true positives, $TN$ the number of true negatives, $FP$ the number of false positives, and $FN$ the number of false negatives.

\subsubsection{F1 score}
\label{ch_03:f1_score}

Also known as the Dice coefficient or overlap index \cite{F1_score}, the F1 score is the harmonic mean of the precision $Pr$ and recall $Re$, which are in turn defined as follows:

\begin{equation}
\label{eq:f1_score_pr}
    Pr = \frac{TP}{TP+FP},
\end{equation}

\begin{equation}
\label{eq_f1_score_re}
    Re = \frac{TP}{TP+FN},
\end{equation}

The F1 score is defined as the harmonic mean of $Pr$ and $Re$ as follows:

\begin{equation}
\label{eq:f1_score}
    F1 = 2*\frac{Pr*Re}{Pr+Re}
\end{equation}

\subsubsection{Hafiane Quality Index}
\label{ch_03:haf}

Proposed by Hafiane et al. \cite{Hafiane07} for fire segmentation evaluation, it measures the overlap between the ground truth and the segmentation results, penalizing as well the over- and under-segmentation \cite{Hafiane07}.

First, the authors define a matching index $M$ as follows \cite{Toulouse15}:

\begin{equation}
\label{eq:hafiane_m}
    M = \frac{1}{Card(I^S)}\sum_{j=1}^{NR^S} \frac{Card(R_{i*}^{GT} \cap R_j^S) \times Card(R_j^S)}{Card(R_{i*}^{GT} \cup R_j^S)},
\end{equation}

where $NR^S$ is the number of connected regions in the segmentation result $I^S$. $R_j^S$ represents one of the said regions, and $R_{i*}^{GT}$ is the region in the reference image $I^{GT}$ that has the most significant overlapping surface with the $R_j^S$ region.

Next, Hafiane et al. define an additional index $\eta$ to take into account the over- and under-segmentation as follows \cite{Toulouse15}:

\begin{equation}
\label{eq:hafiane_eta}
    \eta = 
        \begin{cases}
          NR^{GT}/NR^S       &  \text{if $NR^S \geq NR^{GT}$}  \\
          log(1+NR^S/NR^{GT}) &  \text{otherwise}
        \end{cases}.
\end{equation}

Finally, the Hafiane quality index is defined as follows:

\begin{equation}
\label{eq:hafiane}
    HAF = \frac{M + m \times \eta }{1 + m},
\end{equation}

where $m$ is a weighting factor set to 0.5.

\subsection{Quantitative Results}
\label{subsec:quantitative_results}

Table \ref{tab: Quantitative_results} shows the results obtained by the final implementation of the optimized model in the \ac{FPGA} using \ac{MCC}, \ac{HAF}, and F1 score. It can be observed the pruned model presented a slight drop in performance (3\% in \ac{MCC}) whereas the \ac{FPGA} model presented a slightly higher drop (of about 5\% both in \ac{MCC} and F1 score) of performance for all metrics.

This slight degradation is expected given the heavy optimization undergone by the model when passing from 64-bit to 8-bit data representation. However, the gain in throughput (and thus inference time) is significant: the full precision model runs at 8 FPS in a GPU, consuming a large amount of power, whereas our model can attain up to 33.64 \ac{FPS} in the selected \ac{FPGA} when running in multi-threaded mode (15.77 \ac{FPS} for the single-threaded mode), for a fraction of the power consumption.

\begin{table}[t]
    \centering
    \begin{tabular}{c|c|c|c}
    \cline{1-4}
    Model & \ac{MCC} & F1 score & Hafiane \\
    \hline \hline 
    Proposed Model Original (Validation) & \textbf{0.964} & 0.964 & \textbf{0.946} \\ 
    \hline
    Proposed Model Original (Test) & 0.933 & 0.934 & 0.902 \\ 
    \hline
    Proposed Model Pruned (Validation) & 0.964 & \textbf{0.965} & 0.941 \\ 
    \hline
    Proposed Model Pruned (Test) & 0.924 & 0.926 & 0.877 \\ 
    \hline
    Proposed Model FPGA (Validation) & 0.932 & 0.933 & 0.899 \\ 
    \hline
    Proposed Model FPGA (Test) & 0.912 & 0.915 & 0.870 \\ 
    \hline \hline
    \cline{2-3}
    \cline{2-3}
    \end{tabular}
\caption{Segmentation comparison for the different model implementations.}
\label{tab: Quantitative_results}
\end{table}

Table \ref{tab: Quantitative_results_References} provides a comparison with other models in the literature. A recent and thorough comparison of the state-of-the-art carried out by Ciprián-Sánchez et al. \cite{Jorge} compared different architectures, image types, and loss functions on the Corsican Fire Database. Here, we compared the bests model from this study (by Akhloufi et al. \cite{Akhloufi} with various losses) using the base metrics (i.e., \ac{MCC}, \ac{HAF}, and F1 score). From the table, it can be observed that the original model outperforms this previous work by about 2\% (0.933 \ac{MCC}), whereas the \ac{FPGA} implemented model attains a similar performance to the best configuration obtained by Akhloufi (0.912 vs 0.910 \ac{MCC}), using a much smaller footprint. 

\begin{table}[t]
    \centering
    \begin{tabular}{c|c|c|c}
    \cline{1-4}
    Model & \ac{MCC} & F1 score & Hafiane \\
    \hline \hline
    
    Akhloufi + Dice + \ac{NIR} & 0.910 & 0.915 & 0.890 \\ 
    \hline
    Akhloufi + Focal Tversky + \ac{NIR} & 0.914 & 0.916 & 0.889 \\
    \hline
    Akhloufi + Mixed focal + \ac{NIR} & 0.828 & 0.843 & 0.802 \\
    \hline
    Proposed Model Original (Test) & \textbf{0.933} & \textbf{0.934} & \textbf{0.902} \\ 
    \hline
    Proposed Model \ac{FPGA} (Test) & 0.912 & 0.915 & 0.870 \\

    \hline \hline
    \end{tabular}
\caption{Comparison of the proposed model (full-precision and \ac{FPGA} implementation) with other models in the state-of-the-art.}
\label{tab: Quantitative_results_References}
\end{table}

    

\begin{table}[t]
    \centering
    \setlength{\paperwidth}{3pt}

    \begin{tabular}{|p{80pt}|p{90pt}|p{90pt}|p{90pt}|p{90pt}|}

    \hline

    Image & 
    Example 1 &
    Example 2 &
    Example 3\\
    \hline


    Ground truth & 
    \includegraphics[width=0.3\textwidth, height=24mm, keepaspectratio]{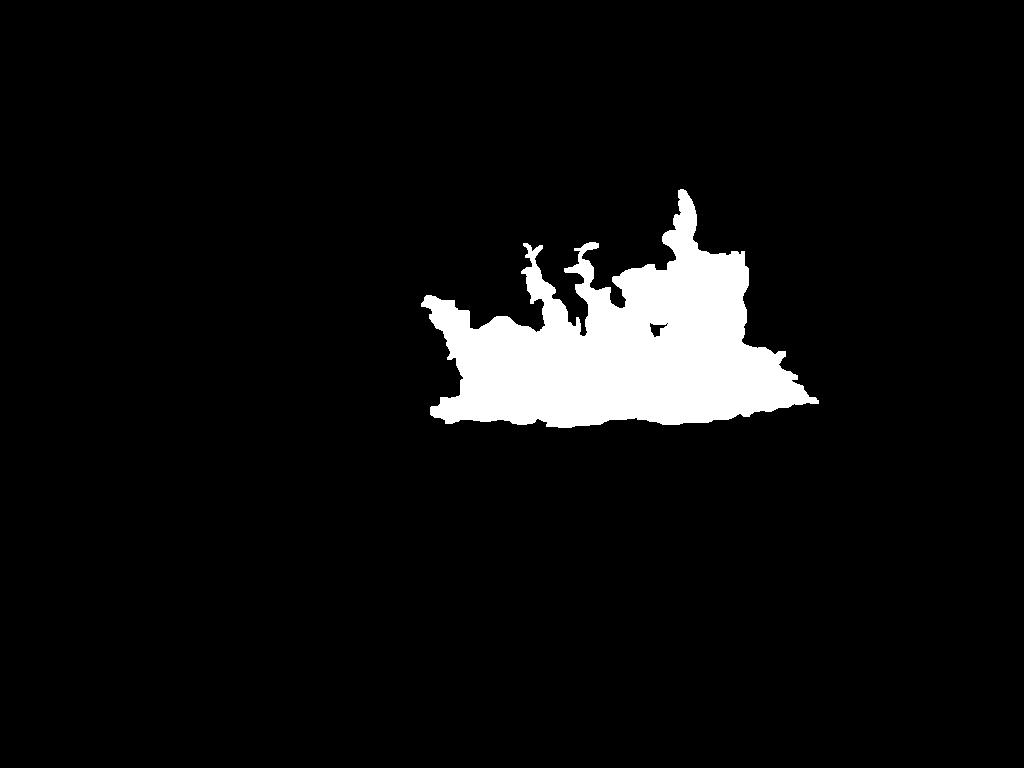} &
    \includegraphics[width=0.3\textwidth, height=24mm, keepaspectratio]{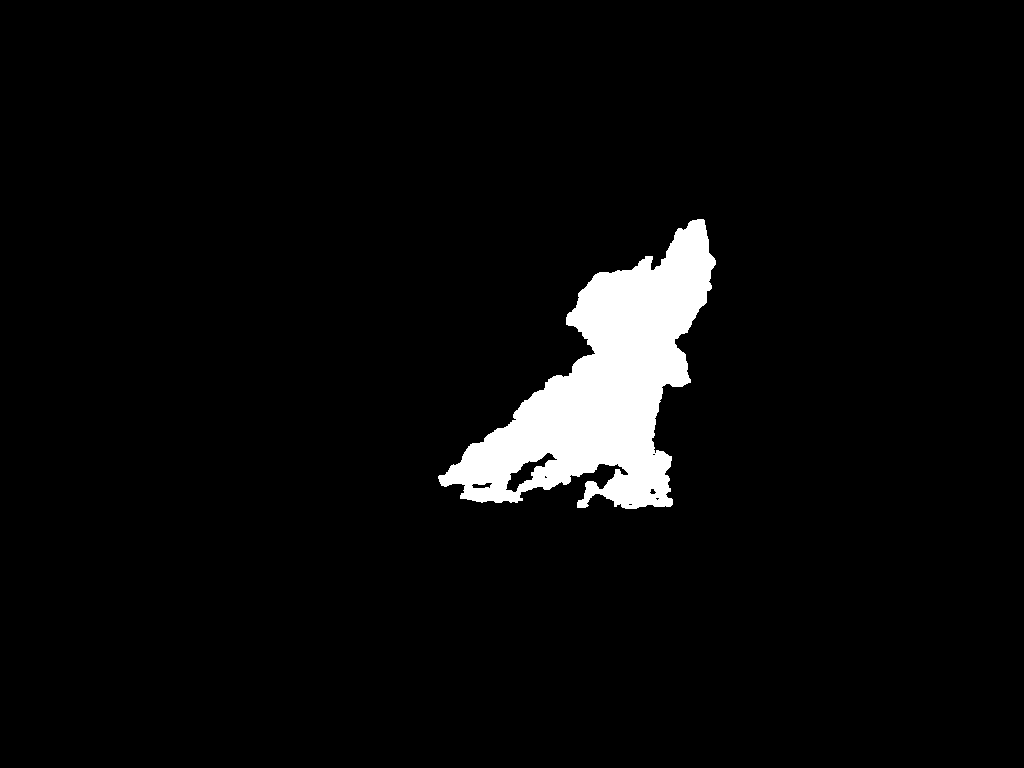} &
    \includegraphics[width=0.3\textwidth, height=24mm, keepaspectratio]{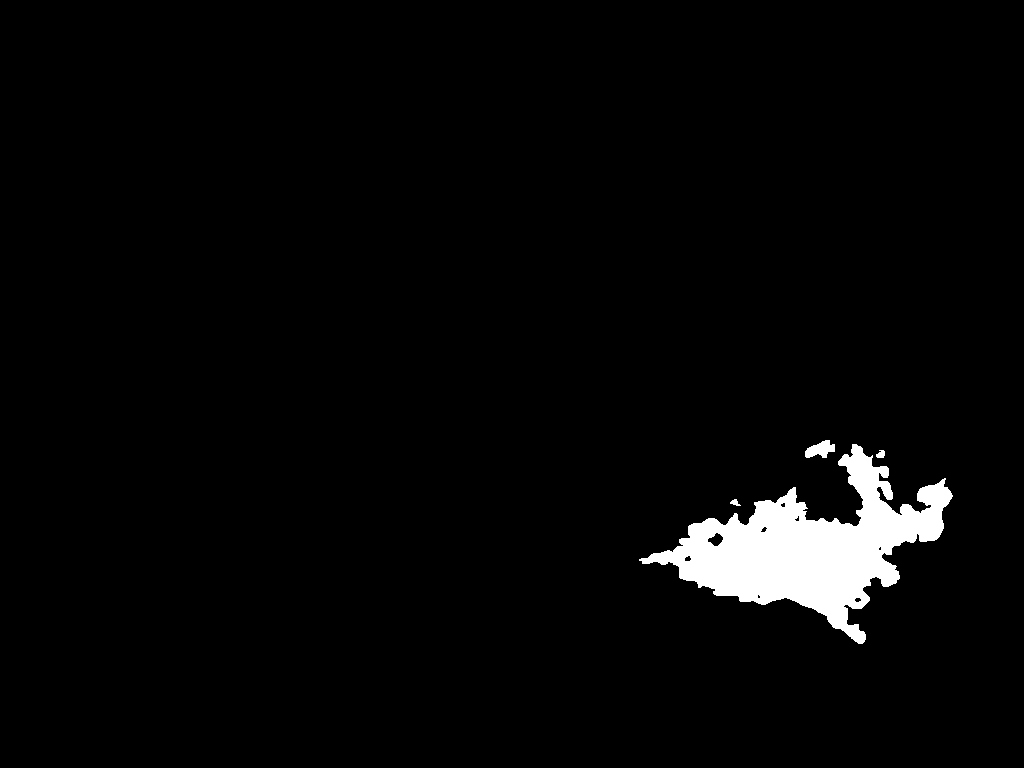} \\
    \hline

    Original model & 
    \includegraphics[width=0.3\textwidth, height=24mm, keepaspectratio]{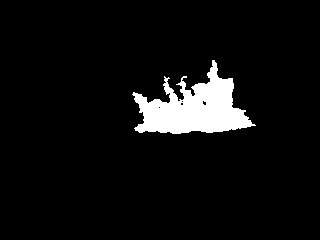} &
    \includegraphics[width=0.3\textwidth, height=24mm, keepaspectratio]{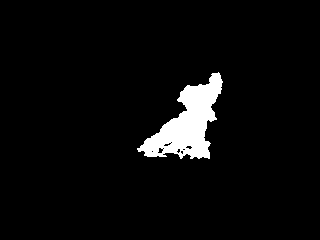} &
    \includegraphics[width=0.3\textwidth, height=24mm, keepaspectratio]{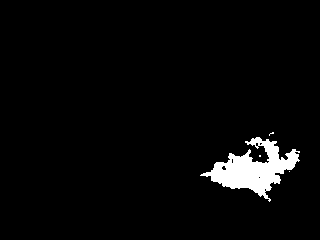} \\
    \hline

    Pruned model & 
    \includegraphics[width=0.3\textwidth, height=24mm, keepaspectratio]{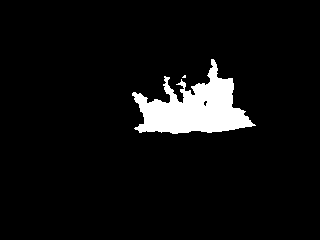} &
    \includegraphics[width=0.3\textwidth, height=24mm, keepaspectratio]{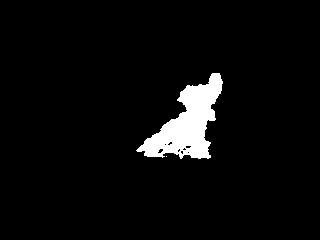} &
    \includegraphics[width=0.3\textwidth, height=24mm, keepaspectratio]{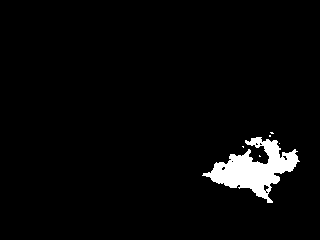} \\
    \hline

    \ac{FPGA} model & 
    \includegraphics[width=0.3\textwidth, height=24mm, keepaspectratio]{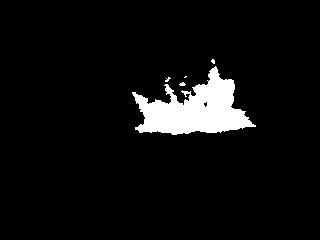} &
    \includegraphics[width=0.3\textwidth, height=24mm, keepaspectratio]{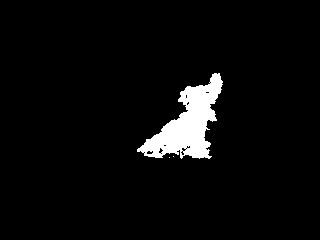} &
    \includegraphics[width=0.3\textwidth, height=24mm, keepaspectratio]{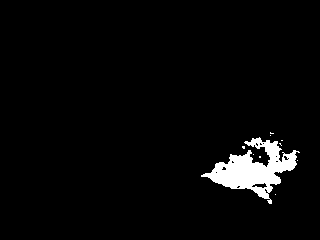} \\
    \hline

    \multicolumn{5}{p{359pt}}{}\\
\end{tabular}
\captionof{table}{Qualitative visual comparison of the segmented images produced by three model configurations: original (full-precision), pruned and quantized (\ac{FPGA} implementation).}
\label{tab: results_visual}
\end{table}

\subsection{Qualitative Results}
\label{subsec:qualitative_results}

Table \ref{tab: results_visual} provides a qualitative comparison of the different models compared in Table \ref{tab: Quantitative_results}. It shows the original images of the Corsican Fire Database and the segmentation results using the original model before the optimization process, after the pruning method, and finally, the final model used in the \ac{FPGA}. It can be observed that for the 3 examples provided, both the pruned model and the \ac{FPGA} implementation yielded practically the same results as the full-precision model, albeit at a much higher frame rate (33 \ac{FPS} vs the 8 the U-Net running on a V100 GPU). Such results can be used in the smart camera for higher image processing tasks in real-time, such as fire spread prediction by using the processing section (ARM processor) of the Ultra96-v2 platform.

\section{Conclusions}
\label{sec:conclusions}

In the present paper, we implement and analyze the performance of a smart camera system based on an \ac{FPGA} accelerator. A modified version of the U-Net architecture was used, to which optimization methods such as quantization and pruning were applied, effectively reducing the inference time and, at the same time, obtaining good results in the wildfire segmentation task. The frame rate obtained in the segmentation task was 33.63 \ac{FPS}. It is believed that there is still some potential to improve the speed of inference by using other strategies, such as the conversion of \ac{CNN} models to \ac{SNN}, whose conversion has been shown to reduce inference times by reducing the number of operations performed \cite{Ju2020}. Finally, given the results obtained, heavy computational tasks are believed to benefit from the accelerators implemented in \acp{FPGA} for their use in real-time applications such as wildfire surveillance using drones.

\section{Acknowledgments}

The authors wish to acknowledge the Mexican Council for Science and Technology (CONACYT) for the support in terms of postgraduate scholarships in this project, and the Data Science Hub at Tecnologico de Monterrey for their support on this project. This work was supported in part by the SEP CONACYT ANUIES ECOS NORD project 315597.
%

%
%
%
\bibliographystyle{plain}
\bibliography{references}
\end{document}